# Analysing Cyberbullying using Natural Language Processing by Understanding Jargon in Social Media

Bhumika Bhatia[1], Anuj Verma[2], Anjum[3], Rahul Katarya[4]

[1,2,3,4] Delhi Technological University, New Delhi, India-110042
bhumika603@gmail.com, ashes4799@gmail.com, anjum_2792@yahoo.com, rahuldtu@gmail.com

**Abstract.** Cyberbullying is of extreme prevalence today. Online-hate comments, toxicity, cyberbullying amongst children and other vulnerable groups are only growing over online classes, and increased access to social platforms, especially post COVID-19. It is paramount to detect and ensure minors' safety across social platforms so that any violence or hate-crime is automatically detected and strict action is taken against it. In our work, we explore binary classification by using a combination of datasets from various social media platforms that cover a wide range of cyberbullying such as sexism, racism, abusive, and hate-speech. We experiment through multiple models such as Bi-LSTM, GloVe, state-of-the-art models like BERT, and apply a unique preprocessing technique by introducing a slang-abusive corpus, achieving a higher precision in comparison to models without slang preprocessing.

## 1 Introduction

One of the upsides, or downsides, of the COVID-19 pandemic, is the ability to solve problems through a digital solution. It has resulted in a lot more online exchange of opinions amongst people with increased social media traffic. This increased exchange of opinions has resulted in a significant increase in cyberbullying worldwide.[1]
Schools, universities, and work-places have all had to adapt to teaching and working in a digitized world through the internet. Cyberbullying can be expressed online through various forms such as racism, sexism, hateful and abusive speech. The definition of cyberbullying varies across different regions, ages, and cultures.[2]What may be socially acceptable in one place may not be considered the same in a different setting. To analyse the effect of cyberbullying across different domains, it is essential to consider the different types of cyberbullying.
Furthermore, making use of a varied and complete dataset that takes into consideration different platforms and across a wide timeline to reflect a holistic view of the types of comments received[3], is crucial. Other challenges faced include segregation of the bullying and non-bullying datasets, which is very subjective and can be prone to human-based annotation errors due to no clear boundary between the two classes, sarcasm, and human-error.
We tackle the task at hand by leveraging **multiple smaller datasets** that have classified different domains of cyberbullying individually and combine those to build a larger dataset. We place emphasis on the importance of building a dataset which can be used to build a **generalized model** that is capable of classifying bullying and normal speech.
Our focus relies on detecting the expression of the language used[4] in online platforms, more specifically, **slang terminology**, which can be difficult to interpret or may be neglected by such models. We handle pre-processing by introducing a novel slang-corpus and removing emojis and links, apart from the general pre-processing techniques followed by [5][6].
In our paper, we introduce a custom-built binary classification model architecture that comprises of Bi-LSTM layers with the use of pre-trained GloVe embeddings trained on a 27B Twitter corpus, as well as compare our performance to other state-of-the-art models such as BERT. Our unique introduction of slang-corpus and text expansion is also shown to improve our precision and accuracy.
In Section 2, we discuss the previous related work in this domain and the research gaps, and then build upon the existing body of work. In Section 3, we address our dataset collection, pre-processing, and the application of various deep neural network model architectures. In Section 4, we have shown our experiment results and graphs, analysing the best model for our evaluation. In Section 5 and 6, we discuss the limitations of our model, concluding the study and also providing a brief overview of future scope in this area.

## 2 Related Work

A survey of related work done to detect cyberbullying using different techniques is presented, which extends the knowledge with the current work done in this area, providing computational motivation, especially, post COVID-19. Scholars and researchers have been struggling with the detection of cyberbullying due to the numerous categories within cyberbullying. With the addition of sarcasm, this task of detection becomes even more challenging.

Researchers used simple classifiers and hard-coded features in 2015-1017. In 2016, Waseem and Hovy[7] used a Logistic Regression model with character level features. In 2017, Davidson used Logistic Regression with word-level features, part-of-speech, sentiment, and some meta-data associated with tweets. These methods were inaccurate, and the user data (features) was not always available.

In 2017-2018[7], to reduce feature engineering overhead, neuronal models were proposed. For example, CNN with character and word level embeddings using logistic regression were applied in binary classification tasks. Park and Fung[8], in 2017, using the fine-tuned embeddings, clustered the vocabulary of the Waseem dataset and found clusters grouped the same classes. In 2018, to overcome the limitation of small datasets, Sharifirad, Jafarpour, and Matwin[9] applied text augmentation and text generation with certain success. Later, research showed that logistic regression classifiers produced better results than lexicon-based classifiers.Karan and Snajder[10](2018) applied a frustratingly easy domain adaptation (FEDA) framework by joining two datasets A and B from different domains. Their features are used and classified with the use of SVM. Later in 2018, pre-trained word embedding with CNNs and GRUs was used, which resulted in improved accuracy.

Celestine Iwendi and Gautam Srivastava[11], in 2020, used double input, output gates, and forget gates on top of the traditional LSTM. Their experiment gave a higher accuracy on the cost of computational complexity. Self-attention models have also been used increasingly, and a review covering all approaches in recent years have been summarised in [12].

# 3 Methodology

In our experimental approach, we divide our work methodology into four parts: Dataset Collection, Pre-processing, Model Architecture and our approach.

## 3.1 Dataset Collection

We collected data from various open-source hate-speech datasets crowd-funded and annotated by several people which include:

- **WASEEM dataset (2016)** collected by Waseem and Hovy [13]consists of 9783 annotated tweets (expert), originally consisting of 16,914 tweets of which the rest have either been deleted or removed as of 2020. It is a manually annotated dataset which includes common racial or sexual slurs, attacks based on religion, or a form of hate speech directed towards a person of minority. The labels include sexism, racism and neither.
- **ENCASE Horizon 2020 hate-speech twitter** is a dataset made available through the paper "Large Scale Crowdsourcing and Characterization of Twitter Abusive Behavior".[5]It is collected through crowd-sourcing abusive behaviour, which has between 5-20 annotators. They have annotated 80k tweets out of which we managed to extract 57,179 tweets. The labels include normal, spam, abusive and hate as labels.
- **Kaggle Formspring data for Cyberbullying Detection** is from a Summer 2010 crawl[14] which includes 12,857 samples with labels as severity of cyberbullying from 0-10.

**Table 1.** Dataset statistics with Binary Classification

| Dataset Source | Cyberbullying | Normal |
|---|---|---|
| WASEEM | 2031 | 7752 |
| ENCASE-H | 20706 | 36473 |
| FORMSPRING | 1124 | 11733 |

| Dataset Source | Cyberbullying | Normal |
|---|---|---|
| TOTAL | 23861 | 55958 |

We do not perform oversampling or under sampling on our dataset to adjust the distribution of classes as we want our dataset to reflect the true percentage of cyberbullying and normal speech found through online sources, where approximately 70% is from normal class, and the remaining 30% is from bullying class. Our dataset is split into 70% training set, 20% for the test set, and the remaining 10% was taken as a validation set to evaluate the performance of the model during training.

To understand the overall word usage throughout our dataset, we decided to use data visualization techniques such as a word cloud as shown below.

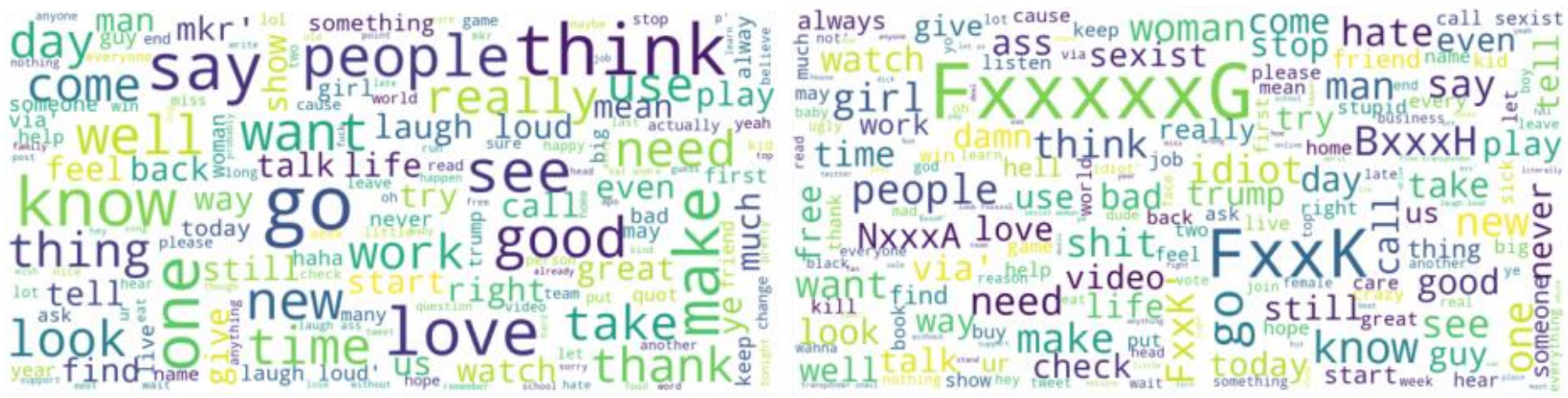

**Fig. 1.** Word Cloud of the normal class and bullying class (Strong words are masked)

### 3.2 Preprocessing

As our data was obtained from different sources (Encase[5], FormSpring[14] and Waseem[13]), the data was very inconsistent and had to be properly pre-processed before it could be put to use for model training purposes. In the resulting dataset, a lot of slang language was found, which led us to the formation of our slang dictionary that converts most slang words into conversational English.

We have elaborated the steps below for preprocessing, by removal of:

- Removal of URLs
- Removal of Emojis
- Removal of mentions of Twitter users
- Removal of mention of retweets
- Removal of Hashtag.
- Removal of Stopwords
- Introduction of a Slang corpus
- Lemmatization of words

### a) Formation Of Slang Corpus

As the dataset contains the use of urban slang lingo[15], which is extremely common among social media platforms, and it is vital to handle it. It has no literal meaning, making it difficult for the model to understand the sentiment and its impact on cyberbullying. Hence, we introduce a corpus which consists of 406 English Slang words and their meaning. The key is to understand the impact of the use of slang terminology on classification of cyberbullying by the use of text expansion using the slang-abusive corpus before training the model. For example:

"mofo" : "m*****f*****r", "SOB" : "son of a b***h"

### b) Lemmatization Of Words

In linguistics, lemmatization is the method of grouping the inflected forms of a word together so that they can be analysed as a single object, defined by the lemma or dictionary form of the word. Lemmatization relies on the proper recognition of the root word and its meaning in a sentence, as well as in the wider sense of that sentence, such as adjacent words. Eg: Crying, Cried, Cry, Cries results into Cry.

## 4 Model Architecture

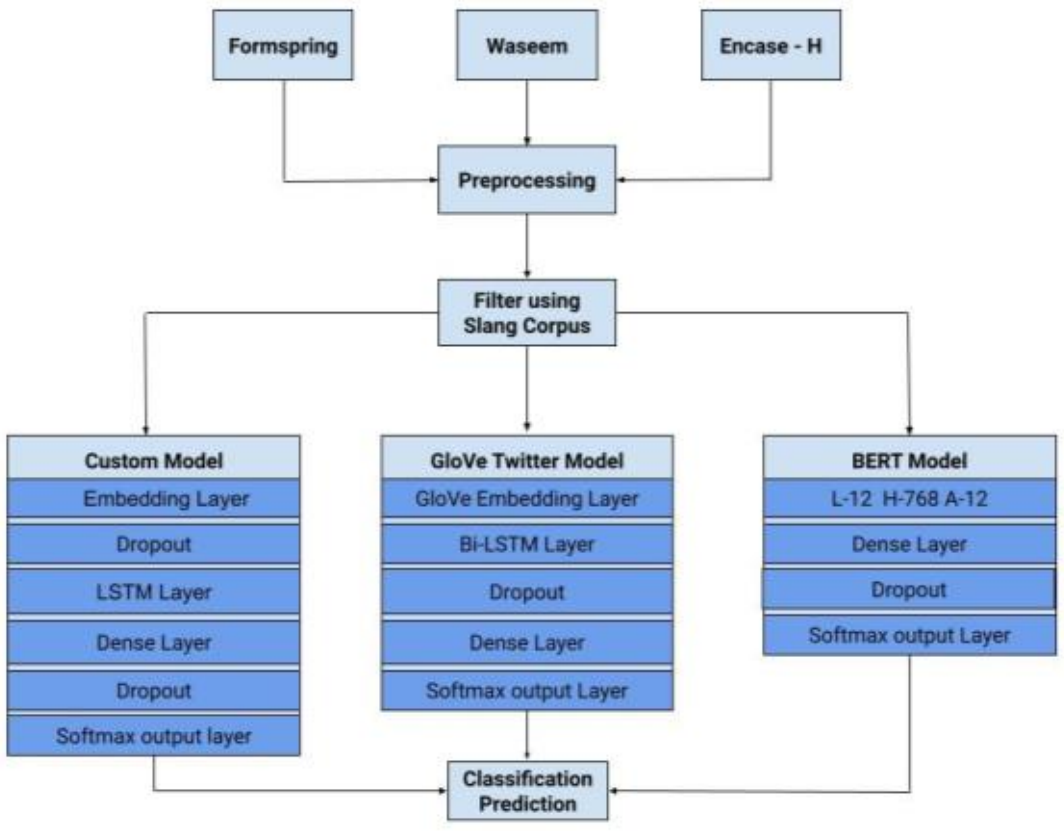

Fig. 2. Flowchart of the methodology

For the detection of cyberbullying, we tried various techniques using different pre-trained embeddings. We also made a custom model. The custom LSTM model does not consist of any pre-trained embeddings, the data is tokenized and then a vocabulary index is formed based on word frequency. The dropout layers used are 0.4 and 0.5. The embedding layer output dimension is 100.

We also decided to experiment with GloVe (Global Vectors for Word Representation) pre-trained word embeddings, using the embedding trained on 2B tweets, 27B tokens and used 100-dimension vectors. We chose to use this particular pre-trained version of GloVe as it is indicative of our training data which is also mostly taken from Twitter. The model's architecture contains a bidirectional-LSTM layer to predict the sentiment of the text with greater accuracy before the output layer. Bidirectional recurrent neural networks (RNN) bring together two separate RNNs.

This arrangement allows the networks to provide backward and forward sequence information at all steps.

$$\hat{y}\langle t\rangle = g(Wy[\vec{a}^{\langle t\rangle}, \overleftarrow{a}^{<t>}]) + b_y \quad (1)$$

$\hat{y}\langle t\rangle$: Output prediction of the layer at time t

Wy : Weight of the layer
by : Bias

Equation (1) specifies the formula for Bi-LSTM. The model is compiled using the nadam optimizer and is trained for 20 epochs.

Lastly, we decided to experiment with Transformer models. Transformer models such as various BERT, XLNet, have seen a lot of success in hate-speech text classification tasks[16]. BERT is Bidirectional Encoder Representations from Transformers. We use the 'bert-base-uncased' model with pre-trained word embeddings. BERT-base consists of 12 layers, 768 hidden layers, and 12 heads for multi headed-attention which captures the broader relationships between words as compared to a single attention mechanism.

This has revolutionized NLP tasks as it is a self-attention model where the entire sequence of words is read simultaneously by the encoder, as opposed to directional models which read the text input sequentially (left-to-right or right-to-left). It is defined as bidirectional, however, describing it as non-directional is more precise. This function allows the model to gain understanding of a word based upon the left and right surrounding word.

It takes input of text in the following format:

- [SEP] - marker for ending of a sentence
- [CLS] - Token added to the start of each sentence as indicator of classification task
- [PAD] - padding for fixed length input to pad shorter sequences
- The rest is encoded by the use of the [UNK] (unknown) token

For the BERT pre-trained model, we modify the end layers by using a dense ReLU layer followed by a dropout of 0.3. Finally, a softmax output layer is used for the binary classification. The results obtained by the following models on the use of slang corpus in the filtering during preprocessing as well as without it on the overall dataset are illustrated in the table below.

## 5 Experiments And Results

After experimenting with our custom, GloVe and BERT model, we have used various evaluation metrics in order to understand and holistically assess our model performance. In addition to the consideration of accuracy, we also look at other metrics such as ROC Curve (Receiver operating characteristic curve), confusion matrix over new data, F1 score, precision and recall. The result for this comparison above are shown in the table below:

**Table 2.** Experiment results for binary classification

| **Dataset** | **Evaluation Metric** | | | | | |
|---|---|---|---|---|---|---|
| | ***Model*** | | ***Precision*** | ***F1 Score*** | ***Accuracy*** | ***AUC*** |
| ENCASE+ Formspring+ WASEEM WITH SLANG | 1. | Custom | **0.84** | 0.69 | 0.84 | 0.84 |
| | 2. | GloVe-twitter | **0.84** | 0.69 | 0.85 | 0.84 |
| | 3. | BERT | **0.67** | **0.72** | **0.84** | **0.88** |
| ENCASE+ Formspring+ WASEEM WITHOUT SLANG | 1. | Custom | 0.76 | 0.69 | 0.84 | 0.84 |
| | 2. | GloVe-twitter | 0.83 | 0.70 | 0.85 | 0.84 |
| | 3. | BERT | 0.66 | 0.70 | 0.83 | 0.87 |

In addition, our use of the slang corpus has effectively proven useful as it has helped improve the F1 score on the combined dataset using BERT model and has shown an increase in precision in all models tested, especially, the custom LSTM model where **precision has improved by 8%**. This is a useful finding as it has reflected a reduction in the false positive rate. This is crucial in terms of application purposes where our goal is to reduce false positives and not wrongly accuse anyone of cyberbullying.

Furthermore, the table below shows the results for ENCASE and WASEEM using the slang pre-processing.

**Table 3.** Experiment results for individual datasets

| **Dataset Source** | **Model** | **Precision** | **F1 Score** | **Accuracy** |
|---|---|---|---|---|
| ENCASE-H | Custom | 0.82 | 0.73 | 0.82 |
| | GloVe-twitter | 0.87 | 0.71 | 0.82 |
| | BERT | 0.87 | 0.82 | 0.88 |
| WASEEM | Custom | 0.86 | 0.75 | 0.91 |
| | GloVe-twitter | 0.86 | 0.75 | 0.91 |
| | BERT | 0.82 | 0.73 | 0.90 |

In comparison to previous work done on datasets, our GloVe model using Bi-LSTM outperforms the previous work result[17] on the Twitter (WASEEM) dataset, giving a higher precision on the positive bullying class.

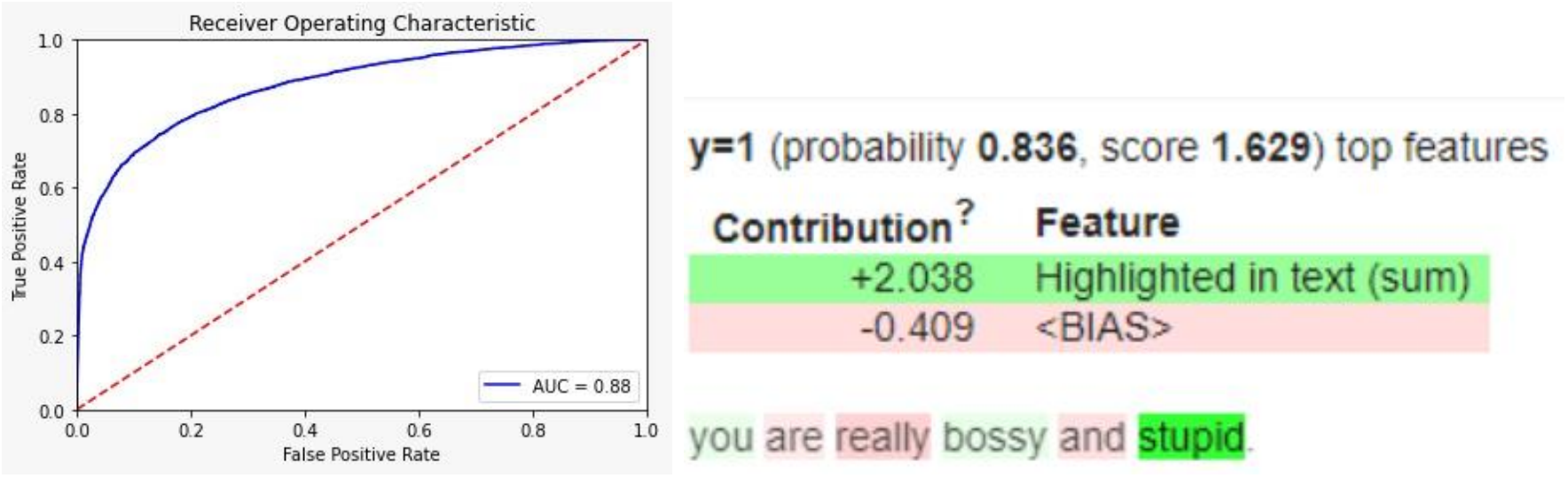

**Fig. 3.** ROC curve for BERT Model on combined dataset and extracting weight measure of individual words in BERT prediction model

We use the ROC curve as a metric as it is a better indication compared to accuracy, which is not the best metric in the case of datasets which are not imbalanced. The BERT model reflects the highest area under the curve of 0.88 over the combined dataset using the slang preprocessing function.

## 6 Limitations

Our work so far is limited to binary classification of cyberbullying. As we are aware, cyberbullying can be further categorized into the categories of sexism, racism, abusive, hate and spam as seen in the Waseem and ENCASE-H hate-speech datasets. The analysis can be further extended to a multi-classification task in order to evaluate how well the models proposed are able to classify within those categories.

In addition, the classification task is very subjective, because the decision on whether something is considered as cyberbullying or not is dependent on the annotator's perspective, the culture, country as well as the age and maturity of the person. This raises questions regarding the authenticity and ability of the annotator to correctly classify the tweets, and in the WASEEM dataset there were found to be some false positives (falsely classified as bullying) and false negatives (falsely classified as normal) within the dataset. This can affect the results of our model and is therefore, a limitation.

Moreover, sarcasm[18] is difficult to detect in text. Since we have not explicitly handled sarcasm, some of the false positives (FP) that we encountered in our model were classified as cyberbullying but were actually normal tweets. Many of these tweets that have cyberbullying that are wrongly classified are those that do not necessarily contain any vulgar or abusive terms that have a strong sentiment towards a particular class.

## 7 Conclusion And Future Scope

Conflating hate material with offensive or innocent words prompts automated cyberbullying detection software to inappropriately mark user-generated content. Not resolving this topic may have significant negative effects such as defaming the users. In order to increase the efficiency of a detection method and generalize it to new datasets, we suggest a transfer learning approach that benefits the pre-trained language model BERT. In addition, our use of slang preprocessing corpus can further be extended by increasing the number of words in the corpus, which currently consists of 406 words.

As our current model is limited to just binary classification of cyberbullying, it doesn't help us to determine the severity of bullying in motion. To further expand on this work, the use of multi classification will be capable of classifying bullying within various classes such as Racism, Sexism, Abusive and Hateful and will also be able to determine the severity of bullying which will be able to create a safe environment with very few false positives.

Furthermore, more increasingly, memes[19] are becoming a popular medium in order to express opinions online and are a newer form of cyberbullying which is significantly more difficult to detect. The future development is to be done in a manner in which the resulting model can analyse the sentiment of a meme and detect any harmful intents.

The dataset is used from limited types of sources and does not represent bullying completely and hence the trained model cannot detect all forms of bullying in all communities. But as previous work done shows, a model cannot work for all communities[20] which perceive bullying differently, we must find some better way of detecting cyber bullying across different communities by introducing diversity of type of language used and regions within our datasets.